\documentclass[letterpaper, 10 pt, conference]{ieeeconf}  

\IEEEoverridecommandlockouts                              

\usepackage{graphicx}
\usepackage{import}
\usepackage{bm}
\usepackage{amsmath} %
\usepackage{amssymb}  %
\usepackage{multirow}
\usepackage{todonotes}
\setuptodonotes{inline}
\usepackage{url}
\usepackage{algorithm}
\usepackage{algpseudocode}

\usepackage{siunitx}
\DeclareSIUnit \Nm{Nm}
\DeclareSIUnit \rpm{rpm}
\DeclareSIUnit \rad{rad}
\usepackage[hidelinks]{hyperref}
\usepackage{cleveref}
\usepackage{comment}
\usepackage{booktabs}
\usepackage{tabularx}
\usepackage{subcaption}
\usepackage{float}

\usepackage[style=ieee, doi=true, isbn=false, url=false,backend=biber]{biblatex} 
\AtEveryBibitem{%
  \clearfield{note}%
  \clearfield{urldate}%
}
\AtBeginBibliography{\small}
\title{\LARGE \bf
Beyond Anthropomorphism: Enhancing Grasping and Eliminating a Degree of Freedom by Fusing the Abduction of Digits Four and Five
}

\author{
Simon Fritsch$^{1}$\textsuperscript{†},
Liam Achenbach$^{1}$\textsuperscript{†},
Riccardo Bianco$^{1}$\textsuperscript{†},
Nicola Irmiger$^{1}$\textsuperscript{†}, 
Gawain Marti$^{1}$\textsuperscript{†},\\
Samuel Visca$^{1}$, 
Chenyu Yang$^{1}$,
Davide Liconti$^{1}$, 
Barnabas Gavin Cangan$^{1}$,\\
Robert Jomar Malate$^{1}$, 
Ronan J. Hinchet$^{1}$,
and Robert K. Katzschmann$^{1}$ 
\thanks{$^{1}$Soft Robotics Lab, IRIS, D-MAVT, ETH Zurich, Switzerland}%
\thanks{† Equal contribution.}
\thanks{{\tt\footnotesize\{\href{mailto:fritschs@ethz.ch}{fritschs}, 
\href{mailto:lachenbach@ethz.ch}{lachenbach}, \href{mailto:rbianco@ethz.ch}{rbianco},
\href{mailto:nirmiger@ethz.ch}{nirmiger},
\href{mailto:gamarti@ethz.ch}{gamarti},
\href{mailto:svisca@ethz.ch}{svisca},
\href{mailto:chenyang@ethz.ch}{chenyang},
\href{mailto:dliconti@ethz.ch}{dliconti},
\href{mailto:bcangan@ethz.ch}{bcangan},
\href{mailto:rmalate@ethz.ch}{rmalate},
\href{mailto:rhinchet@ethz.ch}{rhinchet}, 
\href{mailto:rkk@ethz.ch}{rkk}\}@ethz.ch}}}

\begin{document}
\maketitle
\thispagestyle{empty}
\pagestyle{empty}

\begin{abstract}

This paper presents the SABD hand, a 16-degree-of-freedom (DoF) robotic hand that departs from purely anthropomorphic designs to achieve an expanded grasp envelope, enable manipulation poses beyond human capability, and reduce the required number of actuators. This is achieved by combining the adduction/abduction (Add/Abd) joint of digits four and five into a single joint with a large range of motion. The combined joint increases the workspace of the digits by 400\% and reduces the required DoFs while retaining dexterity. Experimental results demonstrate that the combined Add/Abd joint enables the hand to grasp objects with a side distance of up to 200 mm. Reinforcement learning-based investigations show that the design enables grasping policies that are effective not only for handling larger objects but also for achieving enhanced grasp stability. In teleoperated trials, the hand successfully performed 86\% of attempted grasps on suitable YCB objects, including challenging non-anthropomorphic configurations. These findings validate the design’s ability to enhance grasp stability, flexibility, and dexterous manipulation without added complexity, making it well-suited for a wide range of applications.

\end{abstract}

\section{Introduction}
\subsection{Motivation}
Robust grasping for robotic manipulation is one of the key issues preventing the usage of robots in many applications \cite{science.aat8414}. The difficulty herein can be attributed to both software \cite{zhang2022} and hardware challenges \cite{hardwareissues}. No robotic manipulator has been able to fully match the dexterity, power-to-weight ratio, and exteroception of the human hand \cite{Controzzi2014}. Commercially available solutions, such as robotic grippers \cite{zhao2023}, the \textit{Shadow Robotic Hand} \cite{andrychowicz2020learning}, the \textit{Allegro Hand} \cite{arunachalam2022dexterous} and the \textit{Leap Hand} \cite{shaw2023leap}, tend to be expensive or overly limited in their capabilities. 

This has led many researchers and start-ups to develop their own hardware for manipulation \cite{robotics12010005}. Consequently, no common solution has emerged and there are major differences in the morphologies and kinematics of leading hand designs. For simple pick-and-place tasks, parallel grippers \cite{appius2022raptorrapidaerialpickup, parallel_gripper} and suction-based designs \cite{suctiongripper} are still of great importance. However, such designs are limited in their ability for fine and dexterous in-hand manipulation. 

In contrast, anthropomorphic hands are more dexterous and able to utilize synergies resulting from environments designed for human hands. In turn, they have considerably higher complexity, lower robustness, and lower strength. A large portion of these drawbacks is due to the increased number of degrees of freedom (DoFs) required to mimic the human hand's 21 major DoFs \cite{dof_hand}. Additional DoFs in a hand lead to more failure points, smaller and more fragile joints, and more motors being required. 

Due to these drawbacks, it is pertinent to carefully choose which DoFs to implement in a hand. This paper presents the Super-Abducting hand (SABD hand), which reduces the required number of DoFs by fusing the adduction/abduction (Add/Abd) joint of digits four and five. The lower joint axis of the combined joint also enables a much greater range of motion (ROM), and with it non-anthropomorphic grasp approaches. These enable monomanual grasping of large objects and increased grasp robustness. At the same time, no loss in manipulation capability was observed as a result of the missing DoF. 

\begin{figure}[t]
        \centering
        \resizebox{\linewidth}{!}{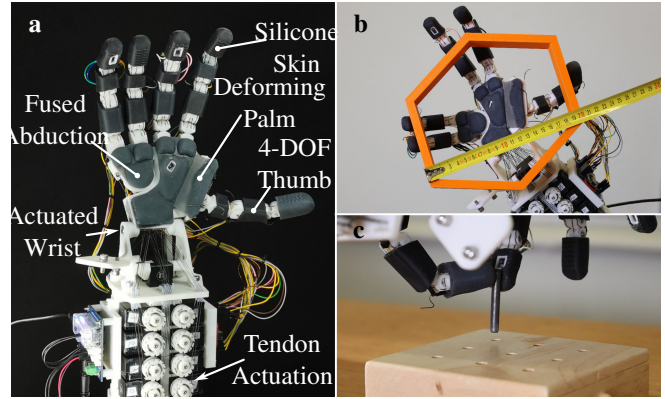}
        \caption{
        Overview of the hand design. (a) Physical implementation of the hand. Select features are labeled. (b) The hand grasping a \qty{200}{\mm} test object using the increased abduction range. Only digits one, four, and five make contact with the test object. (c) The hand performing a peg-in-hole task, demonstrating fine manipulation.}
        \label{fig:Teaser}
\end{figure}

In total, the SABD hand features 16-DoFs including a palm-deforming CMC joint at the thumb and an actuated wrist. An overview of the hand design is shown in \cref{fig:Teaser}.


\subsection{Contribution}
This work presents the SABD hand, an anthropomorphic hand design, which aims to mimic the manipulation capacity of the human hand with a reduced number of DoFs. Additionally, the chosen design enables the hand to utilize non-anthropomorphic grasp approaches. In particular, the key contributions of this work are:

\begin{itemize}
    \item A novel mechanical design for robotic hands using a combined abduction joint for digits four and five. This reduces the required degrees of freedom and enables a much greater range of motion.
    \item A thumb design incorporating a palm joint that enables robust opposition of all fingers and deformation of the palm around grasped objects 
    \item A model-based and experimental evaluation of the hand design, including investigations of the finger workspaces, pinch grasp nullspaces, graspable object size, success rate on YCB objects, and a reinforcement-learning-based grasp robustness measure. 
\end{itemize}
\section{Mechanical Design}
The SABD hand consists of 16 actuated DoFs. All non-standard structural components are FDM 3D printed with a \qty{0.4}{\mm} nozzle. An overview of the hand design can be seen in \cref{fig:Teaser}. 


\subsection{Actuation}

All DoFs in the hand are actuated by servo motors (Robotis Dynamixel XC330-T288-T) with a stall torque of \qty{0.92}{\Nm} and a no-load speed of \qty{65}{\rpm}. During standard operation, the motors are current-limited to \qty{300}{\milli\A}. The wrist is actuated via a 4-bar linkage while all the other DoFs are actuated via tendons. The agonist and antagonist tendons of each joint are wrapped around 2 spools attached to the same motor. The antagonist spool is spring-loaded to account for differences in the displacement of the two tendons. In particular, a torsion spring (Gutekunst Schenkelfeder T-18508R) with a rate of \qty[per-mode = symbol]{1.94}{\milli\Nm\per\degree} is used. Tension of the tendons can be adjusted through the pretension of this spring.  

\begin{figure}[h]
        \centering
        \includegraphics[width=\linewidth]{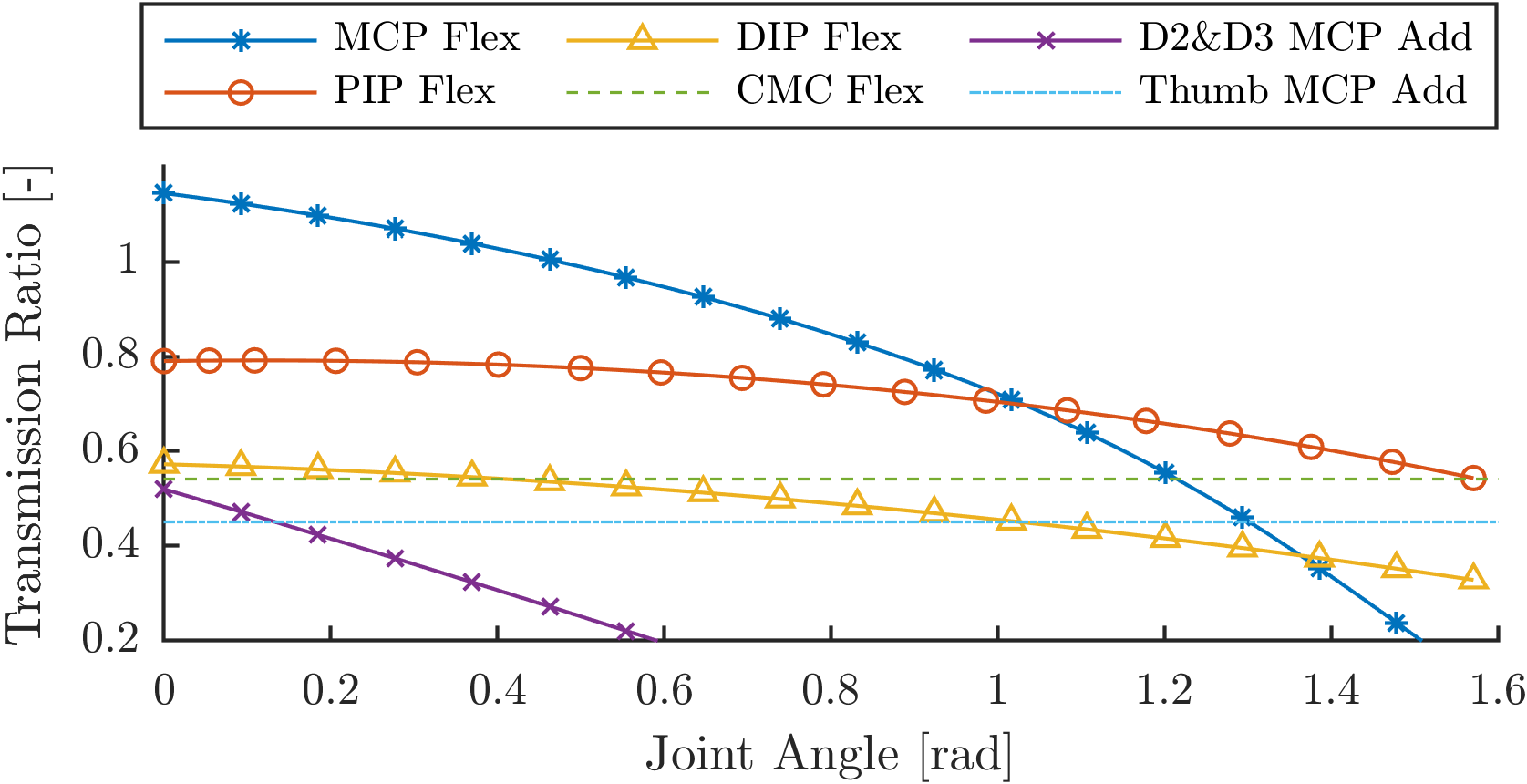}
        \caption{Transmission ratio from the output of the servo motor to the joint as a function of joint angle. }
        \label{fig:TransmissionRatio}
\end{figure}

As a result of the rolling contact finger design, discussed in \cref{finger_design}, the transmission ratios between the motors and the joints are dependent on the angles of the joints. Angle-dependent transmission ratios were approximated by assuming that the tendons do not wrap around the rolling surfaces. The resulting transmission ratios are shown in \cref{fig:TransmissionRatio}.

The motors are mounted on an actuation tower and actively air-cooled. The tendons are routed through PTFE tubes at the top of the tower. From there, they are routed through printed channels in the palm to their respective fingers. At all joints, the tendons that passively cross the joint are routed as close as possible to the center of rotation to reduce unintended coupling. 

\subsection{Design of Digits Two and Three} \label{finger_design}

The design of the fingers was adapted from previous developments in ETH's Soft Robotics Lab \cite{Toshimitsu.2023,hess2024sampling}. Digits two and three have four ligament-stabilized rolling contact joints which will be designated according to the typical convention for human hands \cite{Baker.2023}. 

Starting from the palm, the metacarpophalangeal (MCP) joint allows for flexion/extension (Flex/Ext) and Add/Abd, implemented as two separate rolling contact joints. The proximal interphalangeal (PIP) and distal interphalangeal (DIP) joints each consist of a single rolling contact joint, allowing for movement in the Flex/Ext direction.    

\begin{figure}[h]
        \centering
        \resizebox{\linewidth}{!}{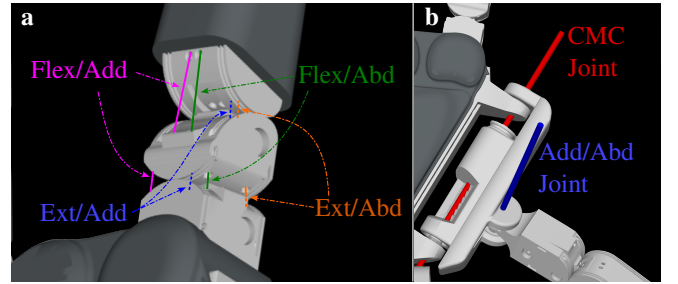}
        \caption{Kinematics of the digits.  
        (a) Tendon arrangement at the MPC joints of digits two and three. Tendons behind the finger are shown as dotted lines. The tendons are labeled with the directions they actuate.
        (b) Axes of the revolute joints of the thumb. Some silicone covers are removed for visual clarity.}
        \label{fig:ThumbKinematics}
\end{figure}

The Flex/Ext and Add/Abd DoFs of each MCP joint are coupled so that one tendon acts on flexion and abduction, one acts on flexion and adduction, and so on. The agonist-antagonist pairings are chosen crosswise so that the flexion-adduction tendon is paired with the extension-abduction tendon. The tendon arragement is shown in \cref{fig:ThumbKinematics}.

In effect, this increases the maximum actuation torque of the MCP joint but comes with a number of drawbacks. The coupling makes control and modeling of the joint more difficult, increases the displacement between the agonist and antagonist, and reduces the maximum speed of combined movements.  

The PIP and DIP joints are rigidly coupled using a tendon that links the extension of the PIP joint with the flexion of the DIP joint. The flexor acts on the flexion of the PIP joint, and the extensor acts on both joints. This arrangement rigidly links the movement of the two joints, but increases the displacement between the flexor and the extensor. 

The distal phalanx design was adapted to house different sensors. In addition, silicone covers with a shore hardness of \qty{10}{A} are added to all phalanges. The covers of the distal phalanges have a ridge pattern to increase grip.

\subsection{Combined Abduction of Digits Four and Five}
The design of digits four and five is similar to that of digits two and three, but the individual rolling contact joints for the MCP Add/Abd motion are replaced by a single revolute joint in the palm. This removes the coupling in the MCP joint and makes MCP Flex/Ext an independent joint. 

The combined abduction joint is located roughly where the distal palmar crease would be on a human hand, centrally in between digits four and five. Its rotation axis is normal to the palm of the hand and actuation of this joint causes the attached palm section to rotate with it. 

The main advantage of this design is a reduction in the required DoFs. We decided to eliminate this DoF in particular because digits four and five are often considered less important than the other digits, as is exemplified by the proliferation of three- and four-fingered robotic hands \cite{Jacobsen.1986,arunachalam2022dexterous,Romero.2024}. Additionally, the MCP joints of digits four and five often abducted together, as exemplified by their correlation coefficient of \num{0.763} in the DLR grasp dataset \cite{Santello.1998}. Another core advantage of this choice is that the placement of these digits allows for a large ROM and with it a non-anthropomorphic grasp strategies. 

In reducing the DoFs, we gain a number of advantages. We save the weight and complexity of an additional motor and reduce the number of tendons we need to route through the hand. Furthermore, we increase the mechanical robustness by eliminating a joint and with it a potential failure point. By placing the joint lower and more central in the hand, we are also able to use a larger diameter axle, which directly increases the robustness of the combined joint. 

\begin{figure}[h]
        \centering
        \resizebox{\linewidth}{!}{
\begingroup%
  \makeatletter%
  \providecommand\color[2][]{%
    \errmessage{(Inkscape) Color is used for the text in Inkscape, but the package 'color.sty' is not loaded}%
    \renewcommand\color[2][]{}%
  }%
  \providecommand\transparent[1]{%
    \errmessage{(Inkscape) Transparency is used (non-zero) for the text in Inkscape, but the package 'transparent.sty' is not loaded}%
    \renewcommand\transparent[1]{}%
  }%
  \providecommand\rotatebox[2]{#2}%
  \newcommand*\fsize{\dimexpr\f@size pt\relax}%
  \newcommand*\lineheight[1]{\fontsize{\fsize}{#1\fsize}\selectfont}%
  \ifx\svgwidth\undefined%
    \setlength{\unitlength}{252bp}%
    \ifx\svgscale\undefined%
      \relax%
    \else%
      \setlength{\unitlength}{\unitlength * \real{\svgscale}}%
    \fi%
  \else%
    \setlength{\unitlength}{\svgwidth}%
  \fi%
  \global\let\svgwidth\undefined%
  \global\let\svgscale\undefined%
  \makeatother%
  \begin{picture}(1,0.35714286)%
    \lineheight{1}%
    \setlength\tabcolsep{0pt}%
    \put(0,0){\includegraphics[width=\unitlength,page=1]{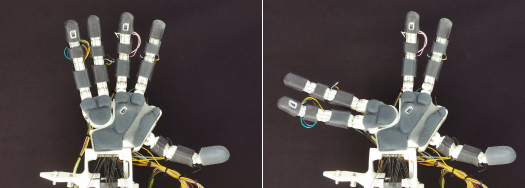}}%
    \put(0.52647877,0.3117373){\color[rgb]{1,1,1}\makebox(0,0)[lt]{\lineheight{1.25}\smash{\begin{tabular}[t]{l}\textbf{b}\end{tabular}}}}%
    \put(0.02743985,0.3117373){\color[rgb]{1,1,1}\makebox(0,0)[lt]{\lineheight{1.25}\smash{\begin{tabular}[t]{l}\textbf{a}\end{tabular}}}}%
  \end{picture}%
\endgroup%
}
        \caption{Increased range of motion of the combined abduction. (a) Base pose of the hand. (b) Fully abducted pose.}
        \label{fig:Superabduction}
\end{figure}

Additionally, this lower placement enables a considerably greater ROM than is usual for MCP Add/Abd in these digits. The hand is able to abduct the attached digits nearly \qty{90}{\deg} which enables configurations such as the one shown in \cref{fig:Superabduction}.

In this configuration, the thumb and digits four and five directly oppose each other while maintaining a large distance between them. This enables the hand to grasp very large objects while maintaining good opposition. 

However, this design also introduces additional complexities. For one, the fusion of the Add/Abd joints means that all tendons for both fingers need to be routed through one joint instead of two. This increases how close to the joint axis they can be routed and as such introduces a stronger unintended coupling. 

Furthermore, the combined joint leads to a branching kinematic chain that needs to be accounted for in the control strategy. This can make the adaptation of control strategies designed for independent finger kinematics more challenging.   

\subsection{Thumb Kinematics}

The IP and MCP flexion joints of the thumb are based on the PIP and MCP flexion joints of the other digits and have identical kinematics. The distal phalanx of the thumb is an elongated version of the distal phalanx of the other digits. 

In addition to these rolling contact joints, the thumb also features a revolute MCP Add/Abd and a revolute one-DoF CMC joint. A closeup of the thumb and its joint axes is shown in \cref{fig:ThumbKinematics}.

This thumb design has two main advantages: It allows the hand to retain a large nullspace during pinch grasps between the thumb and each of the other digits, and it deforms the palm cover around grasped objects.

The nullspace represents the amount of movement the hand is capable of while maintaining a pinch grasp. As such, it is a metric for the capability of the hand to manipulate grasped objects. 

Furthermore, this nullspace can be interpreted as a measure for the robustness of pinch grasps. If the nullspace is large enough, the hand is able to maintain a pinch grasp while compensating for occurring deviations. In turn, a small nullspace represents the fingers just barely being able to touch each other. Consequentially, a hand with a small pinch grasp nullspace is less robust to deviations, and may not be able to achieve a proper pinch grasp if the digits can not achieve their full ROM. 

\begin{figure}[h]
        \centering
        \resizebox{\linewidth}{!}{
\begingroup%
  \makeatletter%
  \providecommand\color[2][]{%
    \errmessage{(Inkscape) Color is used for the text in Inkscape, but the package 'color.sty' is not loaded}%
    \renewcommand\color[2][]{}%
  }%
  \providecommand\transparent[1]{%
    \errmessage{(Inkscape) Transparency is used (non-zero) for the text in Inkscape, but the package 'transparent.sty' is not loaded}%
    \renewcommand\transparent[1]{}%
  }%
  \providecommand\rotatebox[2]{#2}%
  \newcommand*\fsize{\dimexpr\f@size pt\relax}%
  \newcommand*\lineheight[1]{\fontsize{\fsize}{#1\fsize}\selectfont}%
  \ifx\svgwidth\undefined%
    \setlength{\unitlength}{252bp}%
    \ifx\svgscale\undefined%
      \relax%
    \else%
      \setlength{\unitlength}{\unitlength * \real{\svgscale}}%
    \fi%
  \else%
    \setlength{\unitlength}{\svgwidth}%
  \fi%
  \global\let\svgwidth\undefined%
  \global\let\svgscale\undefined%
  \makeatother%
  \begin{picture}(1,0.85714286)%
    \lineheight{1}%
    \setlength\tabcolsep{0pt}%
    \put(0,0){\includegraphics[width=\unitlength,page=1]{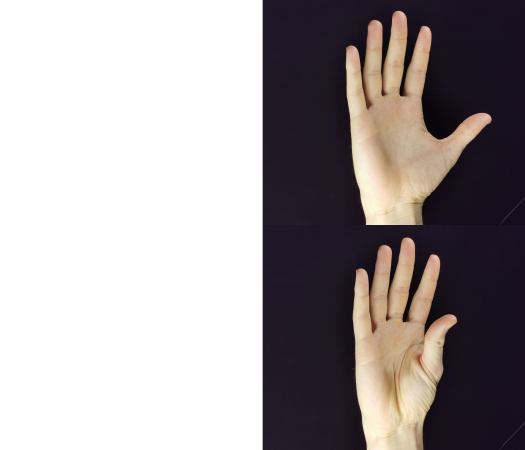}}%
    \put(0.52647877,0.8117373){\color[rgb]{1,1,1}\makebox(0,0)[lt]{\lineheight{1.25}\smash{\begin{tabular}[t]{l}\textbf{b}\end{tabular}}}}%
    \put(0.52726934,0.38316586){\color[rgb]{1,1,1}\makebox(0,0)[lt]{\lineheight{1.25}\smash{\begin{tabular}[t]{l}\textbf{d}\end{tabular}}}}%
    \put(0,0){\includegraphics[width=\unitlength,page=2]{PalmDeformation.pdf}}%
    \put(0.02743985,0.8117373){\color[rgb]{1,1,1}\makebox(0,0)[lt]{\lineheight{1.25}\smash{\begin{tabular}[t]{l}\textbf{a}\end{tabular}}}}%
    \put(0.02725384,0.38316586){\color[rgb]{1,1,1}\makebox(0,0)[lt]{\lineheight{1.25}\smash{\begin{tabular}[t]{l}\textbf{c}\end{tabular}}}}%
    \put(0,0){\includegraphics[width=\unitlength,page=3]{PalmDeformation.pdf}}%
  \end{picture}%
\endgroup%
}
        \caption{Comparison of palm deformation between the SABD hand and a human hand. (a) Base pose of the SABD hand. (b) Base pose of a human hand. (c) Palm deformation of the SABD hand. (d) Palm deformation of a human hand.}
        \label{fig:Palm_Deformation}
\end{figure}

A secondary benefit of the CMC joint is that it deforms the palm around grasped objects. This forms a cup shape that can help locate objects in the palm and may contribute to scoop-type grasps. Such a deformation can also be seen in the human hand. A comparison between the SABD hand's palm deformation and a human hand's palm deformation can be seen in \cref{fig:Palm_Deformation}.

In order to account for this deformation, the silicone pad covering the palm is not rigidly attached to the metacarpal section. It is held in place by two ligament loops around the metacarpal that prevent the pad from bending away from the palm, while maintaining its ability to slide during CMC movement.

\section{Control and Modeling}
The hand is modeled in MuJoCo \cite{MuJoCo}. Each rolling contact joint is implemented as a physical hinge joint coupled to a virtual hinge joint via a tendon. This approach mimics the physical tendon-based design and is adopted from the Faive hand model \cite{Toshimitsu.2023}. 

The digits and wrist are modeled with rolling contact and hinge joints mirroring the real implementation. Each joint of the model is explicitly limited to the ROM of the physical hand. The PIP and DIP Flex/Ext joints are coupled using a tendon-based joint link with a constant coupling factor of \num{0.71}.




To test the capabilities of the hand and gather data for future imitation learning models, a teleoperation setup is used. The operator's hand pose is captured using a wearable motion capture glove (Rokoko Smartgloves)~\cite{caeiro2021systematic}, which is then re-targeted to the joints of the robotic hand. The key challenge herein lies in the morphological and kinematic differences between the human hand and the SABD hand, in particular the increased ROM of the combined abduction joint. 

An energy function-based retargeting strategy~\cite{sivakumar2022robotictelekinesislearningrobotic} was implemented.
This strategy minimizes the difference between key-point vectors defined both on the MuJoCo model and a computationally reconstructed human hand. This optimization-based approach was utilized for the revolute joints of the thumb and the combined abduction of digits four and five. A vector-based angle calculation was implemented for the remaining joints. 

The large Add/Abd ROM in digits four and five required special consideration during the re-targeting process. We introduced a gain for the Add/Abd motion of these digits, which can be adjusted based on the operator's preference and the specific task at hand. This enables the utilization of the increased ROM without compromising performance during human-like grasps.

\section{Results}
In order to assess the performance of the SABD hand, a number of model-based and experimental evaluations were performed. 

\subsection{Workspace and Actuatability}
\begin{figure*}[h]
        \centering
        \resizebox{\textwidth}{!}{
\begingroup%
  \makeatletter%
  \providecommand\color[2][]{%
    \errmessage{(Inkscape) Color is used for the text in Inkscape, but the package 'color.sty' is not loaded}%
    \renewcommand\color[2][]{}%
  }%
  \providecommand\transparent[1]{%
    \errmessage{(Inkscape) Transparency is used (non-zero) for the text in Inkscape, but the package 'transparent.sty' is not loaded}%
    \renewcommand\transparent[1]{}%
  }%
  \providecommand\rotatebox[2]{#2}%
  \newcommand*\fsize{\dimexpr\f@size pt\relax}%
  \newcommand*\lineheight[1]{\fontsize{\fsize}{#1\fsize}\selectfont}%
  \ifx\svgwidth\undefined%
    \setlength{\unitlength}{504.56692913bp}%
    \ifx\svgscale\undefined%
      \relax%
    \else%
      \setlength{\unitlength}{\unitlength * \real{\svgscale}}%
    \fi%
  \else%
    \setlength{\unitlength}{\svgwidth}%
  \fi%
  \global\let\svgwidth\undefined%
  \global\let\svgscale\undefined%
  \makeatother%
  \begin{picture}(1,0.2247191)%
    \lineheight{1}%
    \setlength\tabcolsep{0pt}%
    \put(0,0){\includegraphics[width=\unitlength,page=1]{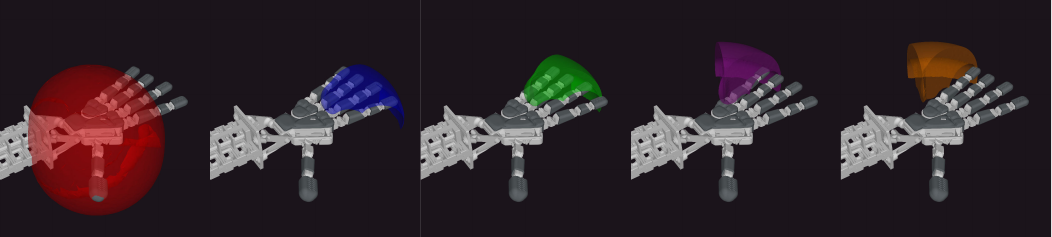}}%
    \put(0.2132231,0.20204183){\color[rgb]{1,1,1}\makebox(0,0)[lt]{\lineheight{1.25}\smash{\begin{tabular}[t]{l}\textbf{b}\end{tabular}}}}%
    \put(0.61361519,0.20204183){\color[rgb]{1,1,1}\makebox(0,0)[lt]{\lineheight{1.25}\smash{\begin{tabular}[t]{l}\textbf{d}\end{tabular}}}}%
    \put(0.81376083,0.20204183){\color[rgb]{1,1,1}\makebox(0,0)[lt]{\lineheight{1.25}\smash{\begin{tabular}[t]{l}\textbf{e}\end{tabular}}}}%
    \put(0.01370451,0.20204183){\color[rgb]{1,1,1}\makebox(0,0)[lt]{\lineheight{1.25}\smash{\begin{tabular}[t]{l}\textbf{a}\end{tabular}}}}%
    \put(0.41360881,0.20204183){\color[rgb]{1,1,1}\makebox(0,0)[lt]{\lineheight{1.25}\smash{\begin{tabular}[t]{l}\textbf{c}\end{tabular}}}}%
    \put(0,0){\includegraphics[width=\unitlength,page=2]{FingerWorkspaces.pdf}}%
  \end{picture}%
\endgroup%
}
        \caption{Workspace analysis. Workspace of (a) digit one, (b) digit two, (c) digit three, (d) digit four, and (e) digit five. The colored volumes represent the locations reachable by a representative point on the tip of the digit.}
        \label{fig:Finger_Workspace}
 \end{figure*}

The workspace of the individual fingers was computed using the established Mujoco model. To do this, point clouds of the fingertip position were computed for \num{5e5} random joint positions. Based on these point clouds, an alpha shape representing the digit's workspace was computed. A visualization of these alpha shapes can be seen in figure \cref{fig:Finger_Workspace}.

Additionally, the workspace of digits four and five with a more physiological abduction ROM of \qty{0.2}{\rad} was calculated and compared to the full workspace of the combined abduction. The increased, non-anthropomorphic ROM of the combined abduction increases the workspace of both digits by 400\%. This calculation does not take the lower center of rotation into account, which further increases the SABD hand's workspace when compared to an anthropomorphic design. 

During the experimental evaluation of the hand it was noticed that the actuatability of thumb Add/Abd joint stopped about \qty{15}{\deg} short of its theoretical ROM in both directions. We suspect that this limitation is caused by parasitic coupling of the tendons that cross the joint. Additionally, the flexion of the thumb CMC joint was negligibly limited by the silicone palm pad. A thinner palm pad may reduce this effect. 


Furthermore, a significant coupling between the combined abduction joint and the joints of digits four and five was noted. This limits the extension capabilities of these joints during full abduction, as the coupling is beyond the compensation range of the spring-loaded spools. Additionally, the large deflection angle of the tendons introduces a lot of friction in this configuration, which reduces the effective strength of the attached digits. A better tendon routing through the combined abduction joint is subject to future work.

\subsection{Opposability}
\begin{figure}[h]
        \centering
        \resizebox{\linewidth}{!}{
\begingroup%
  \makeatletter%
  \providecommand\color[2][]{%
    \errmessage{(Inkscape) Color is used for the text in Inkscape, but the package 'color.sty' is not loaded}%
    \renewcommand\color[2][]{}%
  }%
  \providecommand\transparent[1]{%
    \errmessage{(Inkscape) Transparency is used (non-zero) for the text in Inkscape, but the package 'transparent.sty' is not loaded}%
    \renewcommand\transparent[1]{}%
  }%
  \providecommand\rotatebox[2]{#2}%
  \newcommand*\fsize{\dimexpr\f@size pt\relax}%
  \newcommand*\lineheight[1]{\fontsize{\fsize}{#1\fsize}\selectfont}%
  \ifx\svgwidth\undefined%
    \setlength{\unitlength}{252bp}%
    \ifx\svgscale\undefined%
      \relax%
    \else%
      \setlength{\unitlength}{\unitlength * \real{\svgscale}}%
    \fi%
  \else%
    \setlength{\unitlength}{\svgwidth}%
  \fi%
  \global\let\svgwidth\undefined%
  \global\let\svgscale\undefined%
  \makeatother%
  \begin{picture}(1,0.85714286)%
    \lineheight{1}%
    \setlength\tabcolsep{0pt}%
    \put(0,0){\includegraphics[width=\unitlength,page=1]{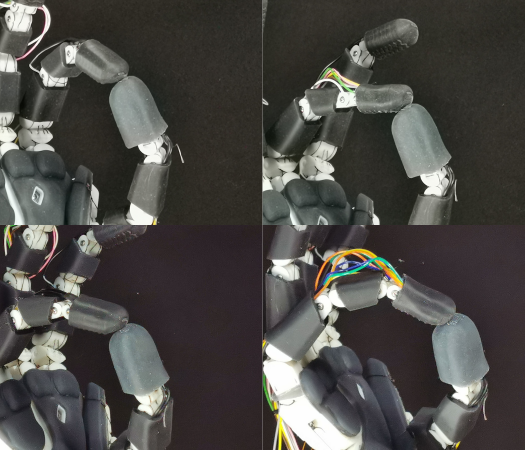}}%
    \put(0.52647877,0.8117373){\color[rgb]{1,1,1}\makebox(0,0)[lt]{\lineheight{1.25}\smash{\begin{tabular}[t]{l}\textbf{b}\end{tabular}}}}%
    \put(0.52726934,0.38316586){\color[rgb]{1,1,1}\makebox(0,0)[lt]{\lineheight{1.25}\smash{\begin{tabular}[t]{l}\textbf{d}\end{tabular}}}}%
    \put(0.02743985,0.8117373){\color[rgb]{0,0,0}\makebox(0,0)[lt]{\lineheight{1.25}\smash{\begin{tabular}[t]{l}\textbf{a}\end{tabular}}}}%
    \put(0.02725384,0.38316586){\color[rgb]{1,1,1}\makebox(0,0)[lt]{\lineheight{1.25}\smash{\begin{tabular}[t]{l}\textbf{c}\end{tabular}}}}%
    \put(0,0){\includegraphics[width=\unitlength,page=2]{ExperimentalPinch.pdf}}%
  \end{picture}%
\endgroup%

        }
        \caption{Thumb opposition tests. Pinching of thumb with (a) digit two, (b) digit three, (c) digit four, and (d) digit five.}
        \label{fig:Experimental_Pinch}
 \end{figure}

Despite the reduced ROM of the MCP Add/Abd joint, the thumb is able to oppose all other digits. Experiments showed the thumb being able to fully achieve a pinch grasp opposition with each of the other digits. Furthermore, after calibration, the hand was able to repeatably and consistently execute these pinch grasps. \Cref{fig:Experimental_Pinch} shows exemplary pinch grasp configurations of the hand. 

\begin{figure}[h]
        \centering
        \resizebox{\linewidth}{!}{
\begingroup%
  \makeatletter%
  \providecommand\color[2][]{%
    \errmessage{(Inkscape) Color is used for the text in Inkscape, but the package 'color.sty' is not loaded}%
    \renewcommand\color[2][]{}%
  }%
  \providecommand\transparent[1]{%
    \errmessage{(Inkscape) Transparency is used (non-zero) for the text in Inkscape, but the package 'transparent.sty' is not loaded}%
    \renewcommand\transparent[1]{}%
  }%
  \providecommand\rotatebox[2]{#2}%
  \newcommand*\fsize{\dimexpr\f@size pt\relax}%
  \newcommand*\lineheight[1]{\fontsize{\fsize}{#1\fsize}\selectfont}%
  \ifx\svgwidth\undefined%
    \setlength{\unitlength}{252bp}%
    \ifx\svgscale\undefined%
      \relax%
    \else%
      \setlength{\unitlength}{\unitlength * \real{\svgscale}}%
    \fi%
  \else%
    \setlength{\unitlength}{\svgwidth}%
  \fi%
  \global\let\svgwidth\undefined%
  \global\let\svgscale\undefined%
  \makeatother%
  \begin{picture}(1,0.85714286)%
    \lineheight{1}%
    \setlength\tabcolsep{0pt}%
    \put(0,0){\includegraphics[width=\unitlength,page=1]{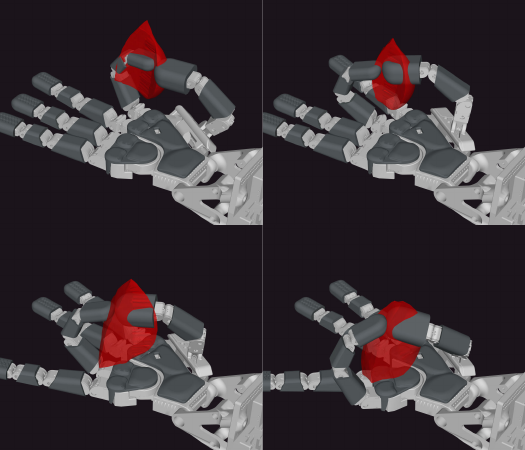}}%
    \put(0.52647877,0.8117373){\color[rgb]{1,1,1}\makebox(0,0)[lt]{\lineheight{1.25}\smash{\begin{tabular}[t]{l}\textbf{b}\end{tabular}}}}%
    \put(0.52726934,0.38316586){\color[rgb]{1,1,1}\makebox(0,0)[lt]{\lineheight{1.25}\smash{\begin{tabular}[t]{l}\textbf{d}\end{tabular}}}}%
    \put(0.02743985,0.8117373){\color[rgb]{1,1,1}\makebox(0,0)[lt]{\lineheight{1.25}\smash{\begin{tabular}[t]{l}\textbf{a}\end{tabular}}}}%
    \put(0.02725384,0.38316586){\color[rgb]{1,1,1}\makebox(0,0)[lt]{\lineheight{1.25}\smash{\begin{tabular}[t]{l}\textbf{c}\end{tabular}}}}%
    \put(0,0){\includegraphics[width=\unitlength,page=2]{PinchNullspace2.pdf}}%
  \end{picture}%
\endgroup%

        }
        \caption{Nullspaces of the different digit combinations while maintaining a pinch grasp. The pinch grasps shown use digit one opposed by (a) digit two, (b) digit three, (c) digit four, and (d) digit five.}
        \label{fig:Pinch_Nullspace}
 \end{figure}

 While maintaining a pinch grasp, the hand still has a considerable actuatable nullspace. An estimate of this nullspace was established by computing the intersection of the thumb workspace with the individual finger workspaces. A visualization of these nullspaces is shown in \cref{fig:Pinch_Nullspace}.

It should be noted that this estimate has a number of limitations. It is based only on single points on the fingertip surfaces and as such does not capture all possible pinch positions. Furthermore, while the nullspace of the hand inherently lives in SE(3), this estimate only captures the translational DoFs. The rotational DoFs of the nullspace are not shown.

\subsection{Grasp Performance during Teleoperation}

\begin{figure*}[h]
        \centering
        \resizebox{\linewidth}{!}{
        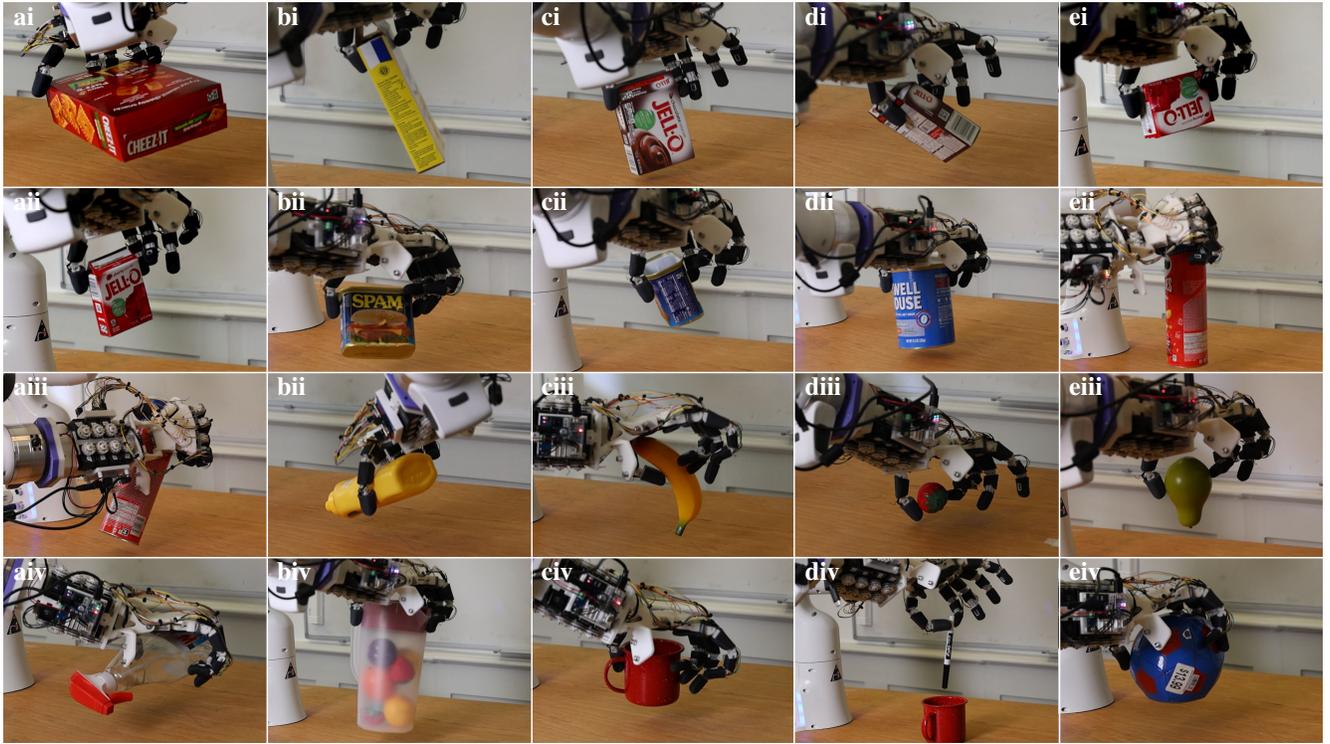
        }
        \caption{Grasps of YCB objects in different orientations. Not all grasped objects and orientations are shown. Shown objects: (ai) cracker box, (bi) sugar box, (ci, di) pudding box, (ei, aii) gelatin box, (bii, cii) potted meat can, (dii) coffee can, (eii, aiii) chips can, (biii) mustard bottle, (ciii) plastic banana, (diii) plastic strawberry, (eiii) plastic pear, (aiv) spray bottle, (biv) pitcher filled with fruit, (civ) mug, (div) small marker from mug, and (eiv) mini soccer ball. Note that the cracker box (ai) and mini soccer ball (eiv) can only be picked up due to the increased abduction ROM. }
        
        \label{fig:Grasp_Experiments}
 \end{figure*}

The grasp performance of the hand was evaluated using teleoperation. For this, objects from the YCB object set \cite{Calli.2015} were selected and the time required to pick them up was measured. Additionally, the success rate of the teleoperated grasps was recorded. The hand was affixed to a Franka Emika Panda robot arm~\cite{haddadin2022franka}  mounted on a table. For each object, six grasp attempts were recorded. At the beginning of each attempt, the arm started in a neutral position above the table, and the object was placed on the table. The orientation of the object was changed for each attempt. Consequentially, two attempts were recorded for every major orientation of cuboid objects. In other words, the boxes were placed twice on every face, up to symmetry. For cylindrical objects, three attempts with the cylinder laying down and three attempts with the cylinder standing up were recorded. \Cref{fig:Grasp_Experiments} shows a selection of the resulting grasps.

After placing the object, a start signal was given and a timer was started. Once the object was lifted off the table and a stable grasp became apparent, the timer was stopped. An attempt was denoted as a failed grasp, if 

\begin{itemize}
    \item The object slipped from the grasp after being lifted up.
    \item The object rolled off the table. 
    \item The arm hit the table causing a safety stop. 
\end{itemize}

Due to the hard, wooden surface of the table, round objects easily rolled during grasp attempts. In particular, the plastic pear and orange were affected by this. Furthermore, three of the selected objects, the plate, bowl, and power drill, were too heavy to be picked up by the hand. The hand was barely able to pick up the weight of the mini soccer ball, which led to unstable and failed grasps. The full pickup times and success rates of all tested objects are denoted in \cref{tab:teleop_grasps}. 

\begin{table}[htbp]
\centering
\caption{Teleoperated Grasping}
\begin{tabular}{l S[table-format=1.2] S[table-format=3.2] }
\toprule
Item & {Time [\unit{s}]} & {Success Rate [\unit{\%}]}  \\
\midrule
Cheez-it Cracker box* & \num{9.31 \pm 5.51} & \num{100} \\
Domino Sugar box & \num{4.48 +- 1.41} & \num{100} \\
Jell-O Chocolate Pudding box & \num{4.52 +- 1.18} & \num{83} \\
Jell-O Strawberry Gelatin box & \num{5.03 +- 1.22} & \num{100} \\
Spam Potted Meat can & \num{4.73 +- 0.80} & \num{100} \\
Coffee can & \num{5.55 +- 0.67} & \num{83} \\
Pringles Chips can & \num{5.98 +- 0.88} & \num{83} \\
French's Mustard bottle & \num{5.90 +- 1.02} & \num{83} \\
Plastic banana & \num{9.40 +- 2.59} & \num{83} \\
Plastic strawberry & \num{7.83 +- 1.48} & \num{100} \\
Plastic pear & \num{25.80 +- 3.87} & \num{50} \\
Plastic orange & \num{11.00 +- 3.20} & \num{67} \\
Windex Spray bottle & \num{8.65 +- 1.75} & \num{100} \\
Scotch Brite Dobie sponge & \num{8.40 +- 1.83} & \num{83} \\
Pitcher filled with plastic fruit & \num{8.46 +- 1.85} & \num{83} \\
Plate & \multicolumn{2}{c}{N/A} \\ 
Bowl & \multicolumn{2}{c}{N/A} \\ 
Wine glass & \num{7.78 +- 1.50} & \num{100} \\
Mug & \num{6.87 +- 1.33} & \num{100} \\
Large marker & \num{10.24 +- 1.85} & \num{83} \\
Small marker in cup & \num{15.72 +- 4.04} & \num{83} \\
Power Drill & \multicolumn{2}{c}{N/A} \\
Mini soccer ball* & \num{13.20 +- 2.43} & \num{50} \\
\bottomrule
\end{tabular}
\vspace{3pt} 

 *: Increased abduction ROM was needed to grasp the object.
\label{tab:teleop_grasps}
\end{table}

It should be noted, that the teleoperator had no prior training with these specific objects. For some of the more challenging objects, they tried out approaches during the setup of the test, but for most objects and orientations, their first attempts, without prior practice were recorded. Furthermore, all attempts were recorded in a single continuous session which may have led to operator exhaustion decreasing performance during later tests. \Cref{tab:teleop_grasps} denotes the objects in the order they were tested.

Overall, the most challenging objects and orientations, aside from those too heavy to be picked up, were: 
\begin{itemize}
    \item The pear, as it had a tendency to roll out of the grasp.
    \item The mini soccer ball, due to its large diameter and high weight.
    \item The cracker box lying flat, due to the large distance between the available parallel sides.     
\end{itemize}

The latter two items were only possible to be picked up using the increased abduction range of digits four and five. Without the increased hand opening resulting from this, the hand would not have been able to grasp two parallel sides of the flat cracker box. 

Additionally, the maximum graspable object size was tested by grasping test objects with increasingly distant parallel sides. Due to the increased abduction ROM and the large workspace of the thumb, the hand was able to grasp the largest test object with a side distance of \qty{200}{\mm}. The resulting grasp is shown in \Cref{fig:Teaser}. An evaluation with larger objects to determine the limit of the graspable object size is still outstanding.

\subsection{Grasp Stability Increase due to the Increased Abduction Range of Motion}

The increase in grasp stability as a result of the increased abduction ROM was investigated using reinforcement learning (RL). The ability of the grasp to resist random forces was used as a stability measure. To this end, separate RL policies were trained with and without using the combined abduction joint. The policy was trained using PPO~\cite{PPO} and the training environment was adapted from faive-gym \cite{Toshimitsu.2023}. 

\begin{figure}[h]
        \centering
        \resizebox{1.06\linewidth}{!}{
\begingroup%
  \makeatletter%
  \providecommand\color[2][]{%
    \errmessage{(Inkscape) Color is used for the text in Inkscape, but the package 'color.sty' is not loaded}%
    \renewcommand\color[2][]{}%
  }%
  \providecommand\transparent[1]{%
    \errmessage{(Inkscape) Transparency is used (non-zero) for the text in Inkscape, but the package 'transparent.sty' is not loaded}%
    \renewcommand\transparent[1]{}%
  }%
  \providecommand\rotatebox[2]{#2}%
  \newcommand*\fsize{\dimexpr\f@size pt\relax}%
  \newcommand*\lineheight[1]{\fontsize{\fsize}{#1\fsize}\selectfont}%
  \ifx\svgwidth\undefined%
    \setlength{\unitlength}{252.00000433bp}%
    \ifx\svgscale\undefined%
      \relax%
    \else%
      \setlength{\unitlength}{\unitlength * \real{\svgscale}}%
    \fi%
  \else%
    \setlength{\unitlength}{\svgwidth}%
  \fi%
  \global\let\svgwidth\undefined%
  \global\let\svgscale\undefined%
  \makeatother%
  \begin{picture}(1,0.44994375)%
    \lineheight{1}%
    \setlength\tabcolsep{0pt}%
    \put(0,0){\includegraphics[width=\unitlength,page=1]{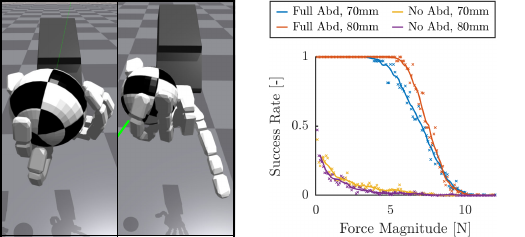}}%
    \put(0.01553509,0.41049058){\color[rgb]{1,1,1}\makebox(0,0)[lt]{\lineheight{1.25}\smash{\begin{tabular}[t]{l}\textbf{a}\end{tabular}}}}%
    \put(0.24538237,0.41049058){\color[rgb]{1,1,1}\makebox(0,0)[lt]{\lineheight{1.25}\smash{\begin{tabular}[t]{l}\textbf{b}\end{tabular}}}}%
    \put(0.47317367,0.41049058){\color[rgb]{0,0,0}\makebox(0,0)[lt]{\lineheight{1.25}\smash{\begin{tabular}[t]{l}\textbf{c}\end{tabular}}}}%
    \put(0,0){\includegraphics[width=\unitlength,page=2]{RLGraspsSuccessRate.pdf}}%
    \put(0.13819531,0.40452406){\color[rgb]{0,0.99607843,0}\makebox(0,0)[lt]{\lineheight{1.25}\smash{\begin{tabular}[t]{l}$F$\end{tabular}}}}%
    \put(0,0){\includegraphics[width=\unitlength,page=3]{RLGraspsSuccessRate.pdf}}%
  \end{picture}%
\endgroup%

        }
        \caption{(a) Exemplary RL-trained grasp of the \qty{100}{\mm} sphere using the combined abduction. The direction of the applied force is highlighted in green. (b) Exemplary RL-trained grasp of the \qty{70}{\mm} sphere without the combined abduction. (c) Ability of the trained policies to resist different random forces.}
        \label{fig:RLRes}
 \end{figure}

As shown in \cref{fig:RLRes}, the hand was placed with its palm facing upwards at a slight angle and spheres of different diameters were chosen as the test objects. For the first second of the simulation, the sphere was pushed into the hand, allowing the controller to establish a solid grasp. After one second, random forces with a magnitude from \qtyrange{0}{5}{\N} were applied to the sphere. These forces were changed after every further second to rigorously assess grasp stability.

While holding the sphere, the agent received a reward of +0.016 at every time step. If the agent dropped the ball, it received a reward of -1. The maximum number of steps for each episode was 600. This straightforward reward structure allows an objective assessment of the abduction mechanism's impact on grasp stability.

Two different setups were tested for the hand with and without utilization of the combined abduction. In setup one, a sphere with a diameter of \qty{70}{\mm} or \qty{80}{\mm} was selected randomly at the start of each episode. In setup two, the sphere was instead selected from \qtylist{70;80;90;100}{\mm}. 

\begin{figure}[h]
        \centering
        \includegraphics[width=\linewidth]{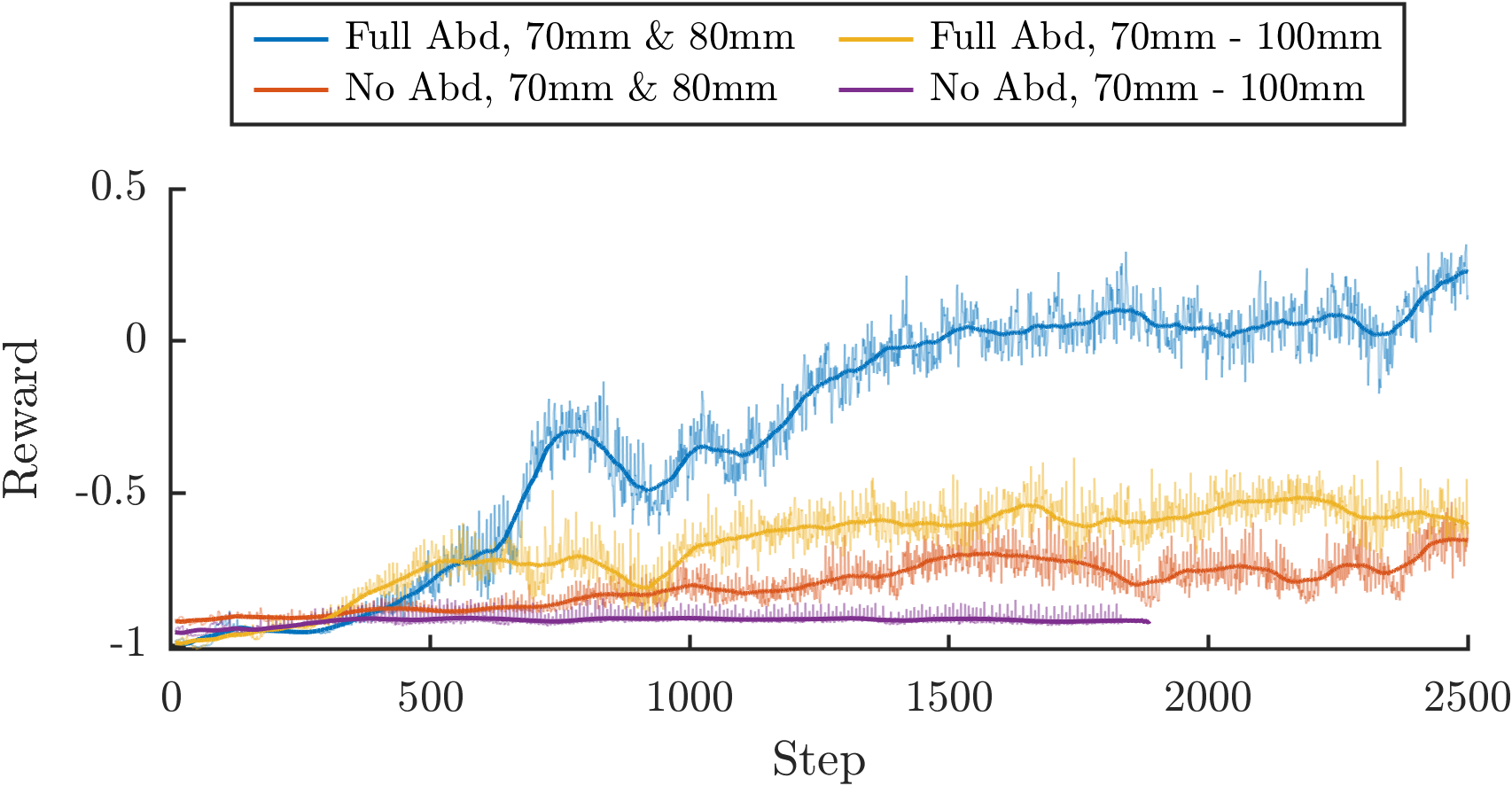}
        \caption{The mean rewards during the training of the grasping task of different RL policies are shown. 
        Policies incorporating the combined abduction (blue and yellow curves) consistently achieve higher rewards than their restricted counterparts across both sphere radius setups. Although larger spheres (indicated by the yellow and purple curves) present a greater challenge than smaller ones, the policies utilizing the combined abduction still secure effective grasps, whereas the restricted policies fail to learn a robust strategy.
        }
        \label{fig:Rewards}
\end{figure}

It was observed that learning is more difficult in setup two, as seen in the reward curves in \cref{fig:Rewards}. The agent with the combined abduction enabled outperformed the agent without the combined abduction feature in both settings. 

To quantify the grasping performance, simulated grasps were disturbed with external forces of different magnitudes, and the success rate of the policies was measured. A grasp was counted as unsuccessful if the force was able to move the sphere out of the hand. A simulation-based assessment was chosen to decouple the results from implementation-specific properties such as tendon routing, sensing, and joint control strategies. 

In all settings and across the whole force range, the hand with the combined abduction enabled performed better than the hand without it. Without the combined abduction, the hand struggled to prevent the sphere from slipping out opposite the thumb. This problem is fixed through the increased abduction ROM and exemplifies the increase in grasp stability resulting from the combined abduction design. \Cref{fig:RLRes} shows exemplary grasps of the trained controllers. Additionally, the success rate at different force magnitudes is shown.

\section{Conclusion}

This paper introduced the SABD hand, a 16-DoF anthropomorphic robotic hand featuring a combined abduction joint with a large ROM. This design eliminated a DoF compared to two independent Add/Abd joints and enabled non-anthropomorphic grasp approaches. Experimental evaluations demonstrated that the combined abduction did not compromise dexterity while enabling the grasping of very large objects. The thumb design proved effective in fully opposing all other digits, contributing to robust and adaptable grasps.

However, certain implementation challenges, particularly in tendon routing, limited the full exploitation of the extended abduction range. The absence of proprioception reduced motion repeatability, while the chosen tendon routing led to nonlinear joint coupling and increased friction-induced unpredictability near the joint limits. Integrating joint sensing could improve control robustness and precision.


Additionally, the benefits of the expanded abduction range were constrained by human-influenced grasp strategies during teleoperation. Prolonged operator training could lead to more effective utilization and further reinforcement learning experiments may better quantify the performance gains enabled by the extended range of motion.


Overall, the combined abduction mechanism presents a promising compromise between having no abduction in the digits and adding two independent DOFs. Future work should refine tendon routing, incorporate proprioceptive sensing, and explore advanced control strategies to fully realize the potential of this design.



\section*{Acknowledgment}
The authors would like to thank the team behind the \href{http://www.rwr.ethz.ch}{Real-World Robotics} course at ETH~Zurich for their support during this project. Additionally, we are grateful for the funding support through a research collaboration with Armasuisse.


\printbibliography


\end{document}